\documentclass[10pt,twocolumn,letterpaper]{article}

\usepackage{iccv}
\usepackage{graphicx}
\usepackage{amsmath}
\usepackage{amssymb}
\usepackage{booktabs}
\usepackage{bbold}
\usepackage{xcolor}
\usepackage{caption}
\usepackage{multirow}
\usepackage{soul}
\usepackage[title]{appendix}

\usepackage[pagebackref=true,breaklinks=true,letterpaper=true,colorlinks,bookmarks=false]{hyperref}

\iccvfinalcopy

\ificcvfinal\fi

\begin{document}

\title{Learning with Difference Attention for Visually \\ Grounded Self-supervised Representations}

\author{Aishwarya Agarwal, Srikrishna Karanam, and Balaji Vasan Srinivasan\\
Adobe Research, Bengaluru India \\
{\tt \scalebox{.7}{\{aishagar,skaranam,balsrini\}@adobe.com}}
}

\twocolumn[{
\renewcommand\twocolumn[1][]{#1}%
\maketitle
\begin{center}
 \centering
 \captionsetup{type=figure}
 \includegraphics[width=0.95\textwidth]{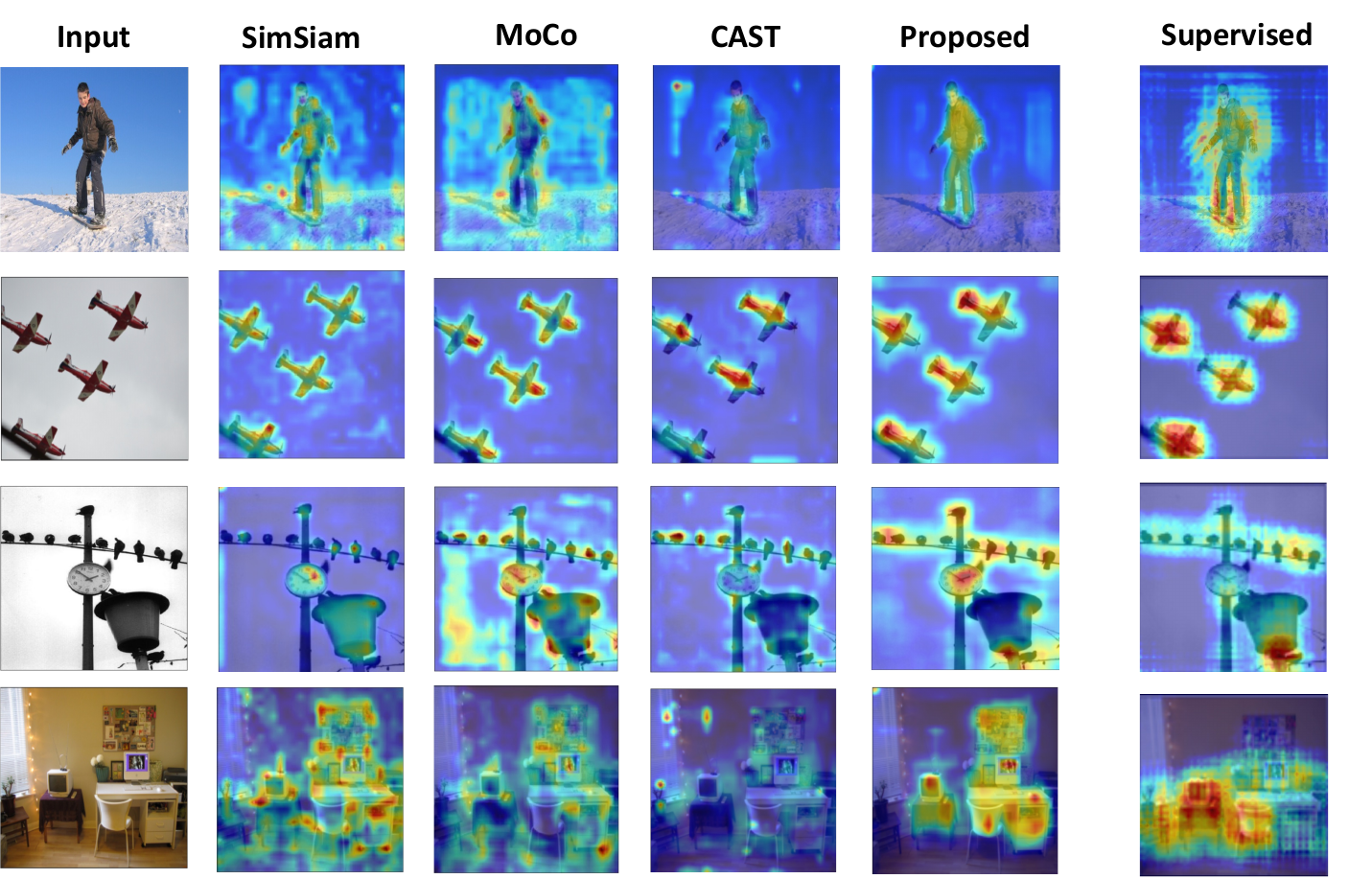}
 \caption{We consider the problem of self-supervised learning (SSL) of image representations and show existing SSL methods are not well grounded (columns 2-4), i.e., they fail to attend to salient image regions (see last column for what a fully supervised model can do). We do this with a new concept called visual difference attention (VDA) that uses feature space differences between an image and its salient-region-masked version. Next, by casting VDA as a differentiable operation, we propose a new loss function that helps improve visual grounding (see ``Proposed") as well as quantitative performance on multiple downstream tasks.
 }

 \label{fig:teaser}
\end{center}
}]

\maketitle
\ificcvfinal\thispagestyle{empty}\fi

\begin{abstract}
   Recent works in self-supervised learning (SSL) have shown impressive results on single-object images, but they struggle to perform well on complex multi-object images as evidenced by their poor visual grounding. To demonstrate this concretely, we propose \textbf{visual difference attention} (VDA) to compute visual attention maps in an unsupervised fashion by comparing an image with its salient-regions-masked-out version. We use VDA to derive attention maps for state-of-the-art SSL methods and show they do not highlight all salient regions in an image accurately, suggesting their inability to learn strong representations for downstream tasks like segmentation. Motivated by these limitations, we cast VDA as a differentiable operation and propose a new learning objective, \textbf{Differentiable Difference Attention} (DiDA) loss, which leads to substantial improvements in an SSL model's visually grounding to an image's salient regions. We first demonstrate this qualitatively with attention maps (Figure~\ref{fig:teaser}) and then show DiDA's quantitative impact with experiments on downstream segmentation, detection, and classification, reporting average precision improvements of 1.3, 1.1, and 0.8 respectively over the state of the art.
\end{abstract}


\section{Introduction}
\label{sec:intro}

With supervised machine learning algorithms relying on expensive-to-create annotated datasets, there has been an increasing interest in self-supervised learning (SSL). In the SSL paradigm, supervisory signals to train machine learning models come from original data samples themselves, thereby removing the need for any external annotations or supervision sources. In particular, recent advances have been driven by approaches based on contrastive learning where positive pairs can come from a variety of sources, e.g., randomly sampled crops from an image \cite{he2020momentum,selvaraju2021casting} or random image augmentations \cite{noroozi2016unsupervised,chen2020simple,chen2021exploring}. Models trained in this fashion have shown competitive results on detection, classification, and segmentation tasks, in some cases outperforming even their fully supervised counterparts \cite{he2022masked,wei2022masked}.

One challenge that exists in any contrastive learning setup is sampling negative pairs to prevent model collapse. In approaches like momentum contrastive (MoCo) learning \cite{he2020momentum} that rely on random image crops, an implicit assumption is crops sampled from the same image would constitute positive pairs while those from other images would form the negatives. While this works well for images with one object, as shown in Contrastive Attention-Supervised Tuning (CAST) \cite{selvaraju2021casting}, it fails in images with multiple objects or even complex scene images since the crops may contain different objects. Further, CAST showed that these methods suffer from poor visual grounding with Grad-CAM attention maps, suggesting they rely on spurious signals (e.g., background) to satisfy the contrastive loss objective.

\begin{figure}
    \centering
    \includegraphics[width = \linewidth]{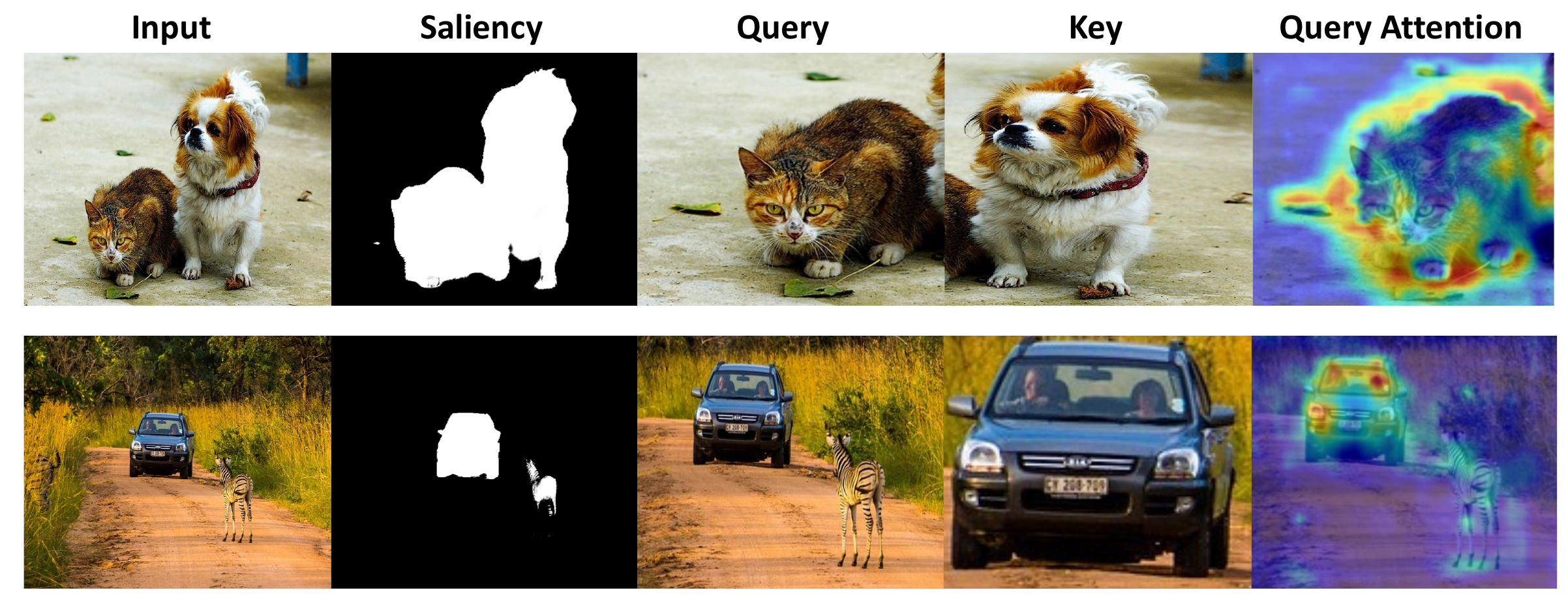}
    \caption{Different-class crops as a positive pair (row 1) and suboptimal attention for images with different-class objects (row 2).
    }
    \label{fig:cast_drawbacks}
\end{figure}

To address the aforementioned issues, CAST \cite{selvaraju2021casting} proposes to use saliency maps precomputed in an unsupervised fashion to constrain the sampling of crops. However, even this would fail on images with multiple objects from various classes. For instance, for an image with two salient objects from two different classes, the saliency random crop operation of CAST \cite{selvaraju2021casting} can return two crops of different classes as a positive pair (see query/cat and key/dog in the first row in Figure \ref{fig:cast_drawbacks}). This is because CAST constrains each crop to cover at least 20\% area of the saliency map and this can result in two crops from different classes being labeled as a positive pair. Using these for contrastive training will naturally lead to undesirable image representations.

To improve visual grounding of SSL models, CAST \cite{selvaraju2021casting} proposes to compute the Grad-CAM attention map with a dot product of feature vectors of two positive-pair crops. Intuitively, this can only capture what is ``common" between the two crops and not all objects in the image. First, given the discussion above, if the crops belong to different classes, this would not result in a meaningful attention map. Second, if the crops have at least one object (from the same class) in common, the attention map would highlight only the common object and not all the objects in the image (e.g., see second row in Figure~\ref{fig:cast_drawbacks} where only car is highlighted in ``Query Attention"). CAST \cite{selvaraju2021casting} uses this attention map as part of an attention loss where the ground-truth is the full-image saliency map that highlights all salient objects (as opposed to CAST's attention map which highlights only the common objects), leading to suboptimal supervision and model underperformance. 

The limitations of CAST \cite{selvaraju2021casting} identified above lead to the principal question we seek to address in the paper- \textsl{how can we equip SSL models with the ability to learn representations of all salient image regions so they can be strong priors for tasks such as multi-class detection and segmentation?} To demonstrate existing SSL methods such as MoCo \cite{he2020momentum}, SimSiam \cite{chen2020simple}, and CAST \cite{selvaraju2021casting} are unable to capture all salient image regions, our first contribution is the introduction of a new concept called \textbf{visual difference attention} (VDA). Our key insight with VDA is that a well-trained model will be able to ``attend to" all salient regions in an image. Masking out all such salient regions and computing a \textsl{difference attention} map between the original image and the masked out image should tell us how well the model has learned these regions. We illustrate this with some example results using  MoCo \cite{he2020momentum}, SimSiam \cite{chen2020simple}, and CAST \cite{selvaraju2021casting}, as well as a fully supervised model (labeled ``Supervised"; we use DeepLabv3 \cite{chen2017deeplab,chen2018encoder}) in Figure~\ref{fig:teaser}. 

From Figure~\ref{fig:teaser}, one can note the supervised model is able to attend well to salient image regions. This is expected since this model was trained in a fully supervised fashion and has learned a good scene prior. This also illustrates VDA's faithfulness in capturing a model's ability to attend to salient image regions. To further substantiate this point, see the results of CAST \cite{selvaraju2021casting}. In the first row, as expected, CAST is able to highlight the person well since there is only one salient object region here. Similarly, in the second row, CAST is able to highlight most aircraft reasonably well (recall CAST discussion above). Note that for these cases, even other baselines \cite{chen2020simple,he2020momentum} work reasonably well because these cases contain objects all belonging to the same class. However, if we consider the examples of rows 3 and 4 that have objects from multiple different classes, CAST is not able to capture all the salient regions. On the other hand, the fully supervised result for these two examples are along expected lines. This discussion shows that while our proposed VDA concept faithfully captures the model's ability to attend to salient regions, in multi-object scenarios, it is inherently the limitations (discussed in paragraphs above) of techniques like CAST \cite{selvaraju2021casting} that lead to poor grounding. Consequently, we introduce a new loss function to address these limitations, leading to the substantially improved results under ``Proposed" in Figure~\ref{fig:teaser}.

To take a step towards SSL models that produce attention maps like the fully supervised ones, and consequently learn stronger scene priors, our second contribution builds on top of the VDA concept. We cast VDA as a differentiable operation that relies solely on activations produced by intermediate layers of the convolutional neural network (CNN) backbone, leading to our new learning objective called \textbf{differentiable difference attention} (DiDA) loss. As discussed above, since CAST \cite{selvaraju2021casting} uses crop attention maps, it is unable to highlight all salient regions since the crop may not contain all regions from the original image. On the other hand, with VDA, we are able to compute the model importance for all salient image regions directly and by training models with the DiDA loss, we explitly force attention maps to be close to all salient regions. We show some examples in column 6 (titled ``Proposed") in Figure~\ref{fig:teaser}, where one can note substantial improvement in visual grounding. In particular, our results highlight all salient regions in the image as opposed to CAST \cite{selvaraju2021casting}, addressing the key limitations identified above. Further, our results are close to the fully supervised oracle, suggesting its potential in learning strong representations for downstream tasks. To this end, given a model trained with DiDA, we conduct extensive experiments on multiple downstream tasks including classification, detection, and instance segmentation, and show substantial improvements over the SSL state of the art. 
Our main contributions are summarized below:

\begin{itemize}
\item We propose a new gradient-based visual attention computation mechanism called \textbf{Visual Difference Attention} to highlight all the image salient regions SSL models attend to, and show they are not visually well grounded, leading to a gap in the current SSL literature.

\item To address the gap above, we cast our proposed VDA concept above as a fully differentiable operation, leading to a new learning objective called the \textbf{differentiable difference attention} (DiDA) loss. By training models with the proposed DiDA loss, we demonstrate substantial performance improvements with the resulting VDA maps now being able to highlight all the salient regions in an image, leading to a strong prior for a variety of downstream tasks/applications.

\item We benchmark the quantitative impact of the DiDA loss with multiple downstream tasks: instance segmentation, detection, and classification and report a new state of the art with average AP improvements of 1.3, 1.1, and 0.8 respectively.
\end{itemize}

\section{Related Work} 

\noindent \textbf{Self-supervised Learning.} SSL methods have typically used the notion of pretext tasks to mine supervisory signals from unlabeled data, including clustering \cite{coates2012learning,tian2014learning,yang2016joint,caron2018deep,caron2019unsupervised,caron2020unsupervised}, and reconstruction and restoration \cite{pathak2016context,zhang2017split}. Much recent effort has also been expended around contrastive learning and their variants \cite{hadsell2006dimensionality} with the resulting models' performance shown to be at par or even outperforming their supervised counterparts \cite{he2022masked,wei2022masked}. While He et al. \cite{he2020momentum} used random crops from an image as positive pairs for contrastive learning, Chen and He \cite{chen2021exploring} used random augmentations of an image to generate these pairs. Selvaraju et al. \cite{selvaraju2021casting} built on top of these ideas to propose an attention-driven learning objective to improve SSL models' visual grounding. One key issue with these methods, as discussed in Section~\ref{sec:intro} and Figure~\ref{fig:teaser}, is their inability to visually attend to all salient regions in an image. We demonstrate this aspect with our proposed new concept of visual difference attention while also showing this operation can be cast in a fully differentiable manner. This results in a new learning objective that improves the model's visual grounding while also providing a strong scene prior for a variety of downstream tasks.  \\
\textbf{Visual Attention Learning.} Since the work of Zhou et al. \cite{zhou2016learning} and Selvaraju et al. \cite{selvaraju2017grad} in computing attention maps for fully supervised classification models, there has been much progress in gradient-based attention learning for convolutional neural networks. In addition to new techniques for computing visual attention maps \cite{chattopadhay2018grad,wang2020score,jiang2021layercam}, much effort has also been expended in extending gradient-based attention computation to other kinds of models, e.g., metric learning \cite{chen2020adapting,ijcai2022p241} and generative modeling \cite{liu2020towards}. Selvaraju et al. \cite{selvaraju2021casting} proposed to extend this line of work to self-supervised learning, using GradCAM \cite{selvaraju2017grad} attention maps to improve the visual grounding of SSL models. However, as discussed in Section~\ref{sec:intro}, since this method used crops from an image as part of the attention loss, the resulting attention map did not attend to all the salient regions in the image. We address this with our DiDA loss, and show training models with this loss results in improved attention maps that attend to all the salient image regions while also giving substantial downstream quantitative improvements.  \\

\section{Approach}
\label{sec:approach}

\subsection{Visual Difference Attention}
\label{sec:vda}

The first contribution of this paper is a method to visually explain, with visual attention maps, the representations learned by existing state-of-the-art SSL models. Using this method, we show the visual grounding capability of existing methods \cite{he2020momentum,chen2020simple,selvaraju2021casting} is far from desirable. While there is much prior work in computing attention maps, e.g., GradCAM \cite{selvaraju2017grad} and extensions \cite{chattopadhay2018grad} for classification models, similarity attention for metric learning models \cite{chen2020adapting,ijcai2022p241} among others, these methods derive their attention signals from available annotation information (e.g., classification logits, pairs of images from same class etc.). In our self-supervised learning scenario, since we do not have access to such labels, these methods cannot be used directly to compute attention maps. Consequently, we propose a new technique to generate these SSL model attention maps.

\begin{figure}
    \centering
    \includegraphics[width = \linewidth]{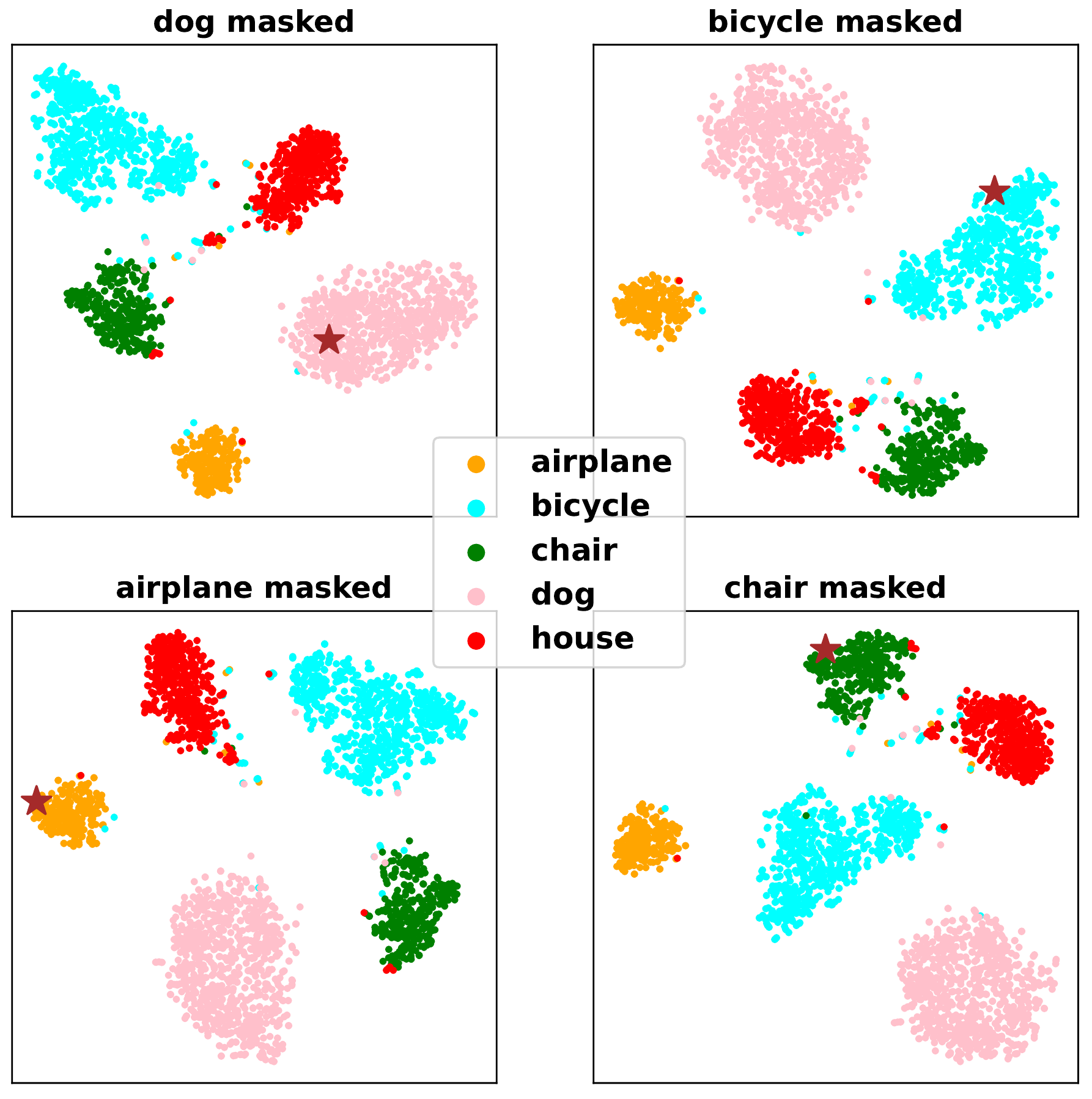}
    \caption{These plots show the difference vector reflects the difference between original and masked images.
    }
    \label{fig:tsne_final}
\end{figure}

Our goal is to show how well an SSL model captures all salient regions, and to this end, our key insight is to compute the importance an SSL model assigns to each salient region in the image when learning image representations. To determine the extent to which the model attends to a certain region in an image, we mask that region out. Let $\textbf{I}$ and $\textbf{I}_{m}$ be the original and the masked images and their feature vectors computed with a model (e.g., output of a fully connected unit) be $\textbf{f} \in \mathbb{R}^{k}$ and $\textbf{f}_{m} \in \mathbb{R}^{k}$ respectively. If the model captured the image region that was masked out well during its learning process, this would be reflected in the difference between the feature vectors $\textbf{f}$ and $\textbf{f}_{m}$ because the only difference between $\textbf{I}$ and $\textbf{I}_{m}$ is the particular region under consideration. In other words, if we take out any image region, this should be reflected in the model output for the modified image and this \textsl{change} measures how well the model attends to this region. 

To substantiate the intuition above, we conduct a toy experiment with some images (about 400) from a random collection of five classes (airplane, bicycle, chair, dog, house) and plot a t-SNE graph (see Figure~\ref{fig:tsne_final}) of their feature vectors computed using DeepLabv3 \cite{chen2017deeplab}. Next, we randomly pick four test scene images (different from the first set) with each having the object of interest. For each test image, we mask out the salient object of interest, compute the $\textbf{f}$ and $\textbf{f}_{m}$ (with DeepLabv3), calculate their difference, and plot this on the t-SNE chart. Our expectation is the following: if we mask out the dog region in the scene image, the resulting difference vector should fall in a cluster of all dog images (and not in any other cluster since the difference between $\textbf{I}$ and $\textbf{I}_{m}$ will be the salient dog region) in the t-SNE chart. This is exactly what happens in Figure~\ref{fig:tsne_final}, e.g., in the ``dog masked" figure, the sample with the asterisk corresponds to the difference vector and we see it is the cluster of points corresponding to the dog class (in pink color). Similar results can be observed in other cases as well.

\begin{figure}
    \centering
    \includegraphics[width = \linewidth]{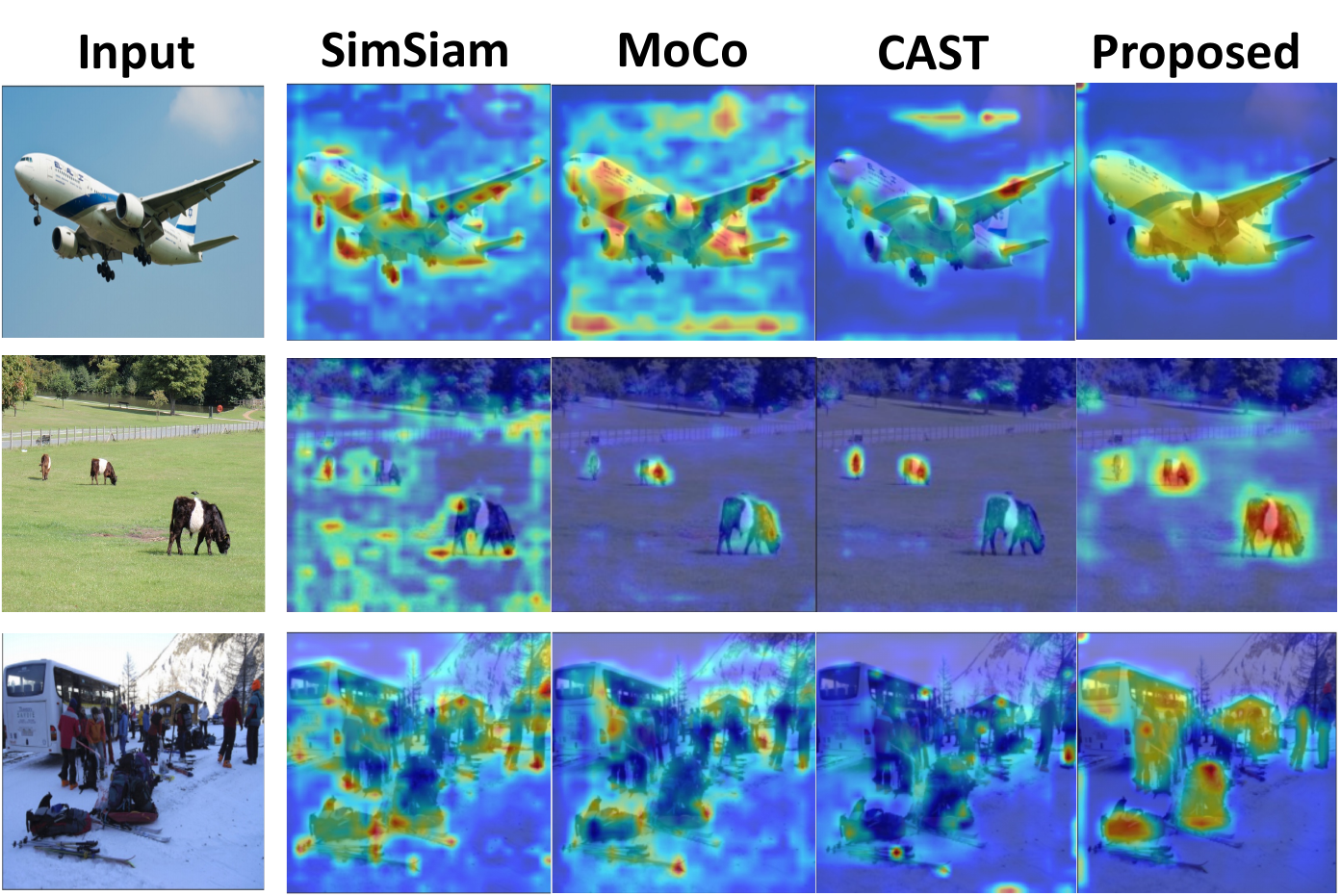}
    \caption{Comparing VDA maps.}
    \label{fig:scene_attention_qual}
\end{figure}

Following the intuition and the experiment above, we conjecture that if we consider the difference feature vector $\textbf{f}_{d}=\textbf{f}-\textbf{f}_{m}$, there will be some dimensions in $\textbf{f}_{d}$ that will contribute the most to capturing the change in the model behavior (i.e., the masked out salient region). Note that we use the saliency maps (computed using DeepUSPS \cite{nguyen2019deepusps} exactly like in CAST \cite{selvaraju2021casting}) associated with each image to determine the region (bounding box) to be masked out. Our idea is to use these feature dimensions to calculate a signal that can then be used to compute a \textsl{difference} attention map called the visual difference attention (VDA). This VDA map will capture the change in the model behavior and reflect the extent to which the model has learned the masked out salient region in its representation. Let $\bar{\textbf{f}}_{d}$ be a vector that retains the elements of $\textbf{f}_{d}$ in the dimensions that contribute the most to this model change, and zeros in all other dimensions. While $\bar{\textbf{f}}_{d}$ can be constructed with
a simple thresholding operation to determine the dominant feature dimensions in $\textbf{f}_{d}$, we show in Section~\ref{sec:dida} this can be cast as a differentiable operation leading to a trainable learning objective. Letting $\mathbb{1} \in \mathbb{R}^{k}$ be a vector of all ones, we first compute the scalar signal $s=\mathbb{1}^{T}\bar{\textbf{f}}_{d}$. We then calculate the derivative of $s$ with respect to the activation maps $\textbf{A} \in \mathbb{R}^{n \times p\times q}$ ($n$ feature maps each of dimensionality $p\times q$) of the last convolutional layer in our encoder, giving the gradient matrix with respect to each feature map $\textbf{G}_i = \dfrac{\partial s}{\partial \textbf{A}_i}, i=1, \cdots, n$. We use an average pooling operation (GAP) on each $\textbf{G}_i$ to obtain the weight $\alpha_i = \text{GAP}(\textbf{G}_i)$. These weights are used to compute the final visual difference attention map $\textbf{M} \in \mathbb{R}^{p\times q}$:

\begin{equation}
    M=\text{ReLU}(\sum_{i=1}^{n}\alpha_i\textbf{A}_i)
    \label{eq:vda}
\end{equation}

High values in the VDA map $\textbf{M}$ indicate the salient regions in the image that were attended to by a model. If the model indeed learned to capture all salient regions in an image, all those regions must have high values in $\textbf{M}$. Since all the discussion above was agnostic of the actual underlying model, we can use the proposed VDA technique to determine how well any SSL model attends to the salient image regions. In Figure~\ref{fig:scene_attention_qual}, following those in Figure~\ref{fig:teaser} in Section~\ref{sec:intro}, we show some sample VDA maps generated using the proposed method with multiple existing SSL methods-  MoCo \cite{he2020momentum}, SimSiam \cite{chen2020simple}, and CAST \cite{selvaraju2021casting}. One can note the visual grounding of these SSL methods is far from desirable and there is much room for improvement (none of these methods highlight the salient image regions accurately). Motivated by these limitations, we next propose a new loss function that fixes these issues.

\begin{figure*}
    \centering
    \includegraphics[width = 1\linewidth]{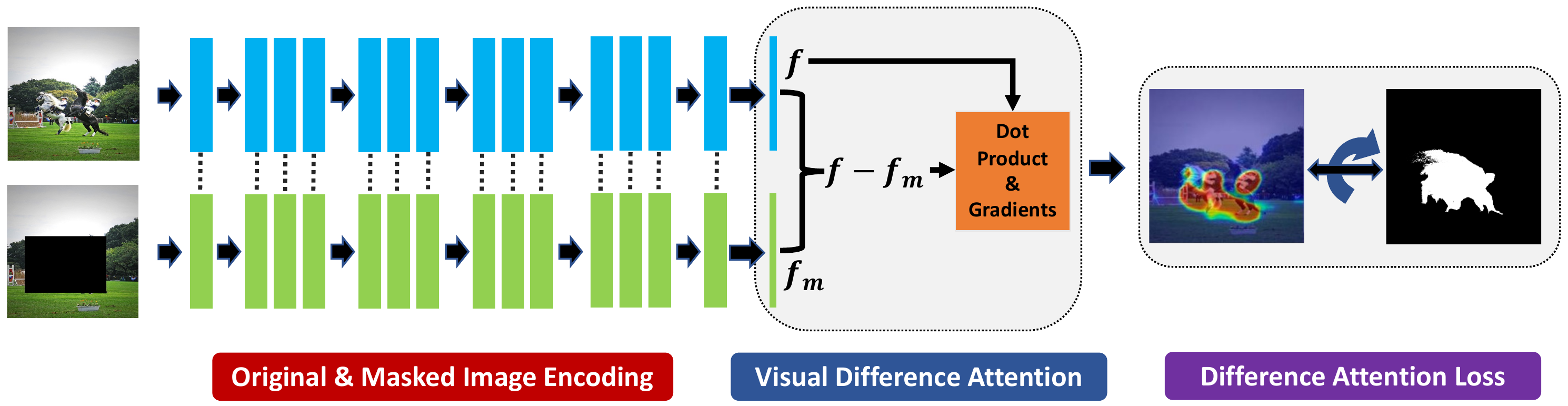}
    \vspace{-18pt}

    \caption{Architecture of our proposed VDA and DiDA Framework.}
    \label{fig:architecture}
\end{figure*}

\begin{figure*}
    \centering
    \includegraphics[width = \linewidth]{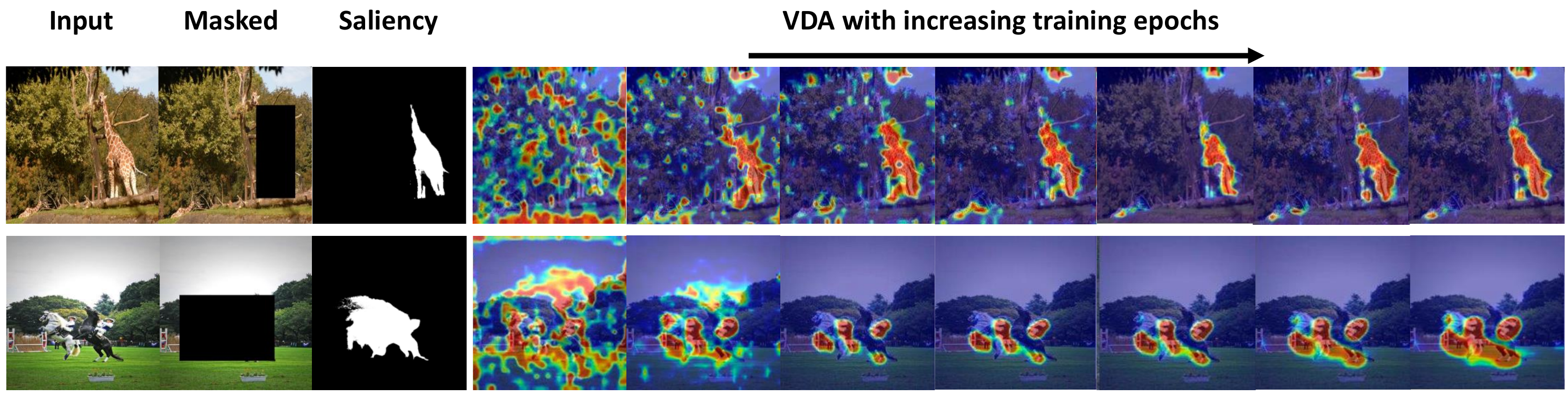}
    \caption{The proposed DiDA loss helps progressively improve visual grounding during training.}
    \label{fig:intermediate_training}
\end{figure*}

\subsection{Differentiable Difference Attention Loss}
\label{sec:dida}

Our goal is to train models so they will be able to attend to all salient regions in an image, thereby improving the results of existing SSL models seen in Figures~\ref{fig:teaser} and~\ref{fig:scene_attention_qual}. As we establish in Section~\ref{sec:results}, this leads to a model that is a better scene prior for multiple downstream tasks such as multi-class classification, detection, and segmentation.  To this end, we use the saliency map from DeepUSPS \cite{nguyen2019deepusps} just like CAST does \cite{selvaraju2021casting}, compare it to the VDA map during training and use the loss to update model parameters (see Figure~\ref{fig:architecture}).

To enable the operation above to be differentiable, we modify the computation of the signal $s$ in Section~\ref{sec:vda} to be differentiable. We achieve this by directly computing the dot product between the feature vectors $\textbf{f}$ and $\textbf{f}_{d}$. Given all vectors to be normalized between 0 and 1, the resulting signal $s=\textbf{f}^{T}\textbf{f}_{d}$ will capture information from feature dimensions corresponding to all salient regions in the original image $\textbf{I}$. This is because $\textbf{f}_{d}$ will have high values in dimensions corresponding to the change between $\textbf{I}$ and $\textbf{I}_{m}$ (that happen to be all masked out salient regions) and multiplying this with $\textbf{f}$ will enhance the corresponding feature dimensions in $\textbf{f}$. This will consequently give a signal that will have focused on all the feature dimensions corresponding to all the salient regions. Given this $s$, our attention computation mechanism remains as in Equation~\ref{eq:vda}, giving the attention map $\textbf{M}$ for the training image $\textbf{I}$. We then use a sigmoid softening operation (helps exclude low response regions and increases importance of the high-response ones) to give the final attention map used in our loss function as $\widetilde{\textbf{M}}=\sigma(\alpha(\textbf{M}-\beta))$ where $\sigma$(.) is the sigmoid function and $\alpha = 16$,  $\beta = 0.5$ are scalars. 
Given $\textbf{I}$'s saliency map $\textbf{S}$, our proposed learning objective is:

\begin{equation}
    \mathcal{L}_\text{DiDA} = 1 - \dfrac{\langle\widetilde{\textbf{M}} , \textbf{S}\rangle}{||\widetilde{\textbf{M}}||||\textbf{S}||}
    \label{eq:dida}
\end{equation}
\vspace{-10pt}

where $\langle\widetilde{\textbf{M}} , \textbf{S}\rangle$ represents the inner product between the flattened versions of $\widetilde{\textbf{M}}$ and $\textbf{S}$, and $||\widetilde{\textbf{M}}||$, $||\textbf{S}||$ are their corresponding Euclidean norms.  

Upon training with this objective, the final attention map $\widetilde{\textbf{M}}$ gets closer to the saliency map $\textbf{S}$, leading to the model attending to all the salient image regions. To demonstrate the impact of $\mathcal{L}_\text{DiDA}$, we show some qualitative results in Figure~\ref{fig:intermediate_training}, where one can note two cases (one in each row). From the fourth column onwards, we show the progression of $\widetilde{\textbf{M}}$ with training epochs. In the beginning, $\widetilde{\textbf{M}}$ does not attend to the correct salient regions, and as training progresses, this improves.

\section{Results}
\label{sec:results}

We conduct extensive experiments to show training a model with our proposed DiDA loss leads to quantitative improvements on several downstream tasks while also resulting in substantial qualitative improvements in the model's visual attention to salient image regions. 

\begin{table*}[h!]
    \centering
    \scalebox{0.75}{
    \begin{tabular}{|c|c|c|c|c|c|c|c|c|c|c|c|}
    \hline
        \multirow{2}{*}{\textbf{Method}} & \textbf{VOC07 clf.} &  \textbf{IN-1k clf.} & \multicolumn{3}{|c|}{\textbf{PASCAL VOC Detection}} & \multicolumn{6}{|c|}{\textbf{COCO Instance Segmentation}}  \\
        \cline{2-12}
        &  mAP  & Top-1 acc. & $AP^{bbox}_{all}$ & $AP^{bbox}_{50}$ & $AP^{bbox}_{75}$ & $AP^{bbox}_{all}$ & $AP^{bbox}_{50}$ & $AP^{bbox}_{75}$ & $AP^{mask}_{all}$ & $AP^{mask}_{50}$ & $AP^{mask}_{75}$   \\
        \hline \hline
        \textbf{Random Init} & – & – &33.8 &60.2 &33.1 &36.7 &56.7 &40.0 &33.7 &53.8 &35.9 \\
        \textbf{ImageNet Fully Sup} & – &– &53.5& 81.3& 59.1 &38.9 &59.6 &42.7 &35.4 &56.5 &38.1\\
        \textbf{COCO Fully Sup} &  86.2 &46.4 &50.9 &79.2 &54.7 &40.3 &61.3 &43.7 &36.5 &58.1 &39.1\\
        \textbf{MoCo-COCO} \cite{he2020momentum} &  67.5 &46.5 &47.5 &75.4 &51.5 &38.3 &58.7 &41.5 &34.9 &55.7 &37.2\\
        \textbf{CAST} \cite{selvaraju2021casting} & 74.0 & 48.7& 54.2& 80.1& 59.9& 39.4& 60.0& 42.8& 35.8& 57.1& 38.6\\
        \textbf{DiDA (Ours)} & \textbf{74.8} & \textbf{49.2}  &  \textbf{54.9} & \textbf{80.6} & \textbf{61.0} & \textbf{40.6} & \textbf{61.1} & \textbf{44.1} & \textbf{37.1} & \textbf{58.2} & \textbf{39.9}\\
        \hline
    \end{tabular}
    }
    \vspace{-0.4em}
   \caption{Evaluation results on downstream tasks.}
    \vspace{-5pt}
    \label{tab: results_downstream_quant}
\end{table*}

\begin{table*}
    \centering
    \scalebox{0.75}{
    \begin{tabular}{|c|c|c|c|c|c|c|c|c|}
    \hline
        \textbf{Method} & \textbf{Original} & \textbf{Mixed-Same} & \textbf{Mixed-Rand} & \textbf{Mixed-Next} & \textbf{Only-FG} & \textbf{No-FG} & \textbf{Only-BG-B} & \textbf{Only-BG-T} \\
        \hline
    MoCo-COCO \cite{he2020momentum} & 72.62 & 45.75 & 30.44 & 26.86 & 30.42 & 23.95 & 5.06 & 12.62 \\ 
     CAST \cite{selvaraju2021casting} &  77.33& 54.42& 39.93& 37.46& 43.26& 23.70& 4.40& 12.59 \\
     DiDA & \textbf{78.91} & \textbf{55.73} & \textbf{40.13} & \textbf{37.82} & \textbf{53.44} & \textbf{24.36} & \textbf{5.53} & \textbf{14.21}\\
     \hline
    \end{tabular}
    }
    \caption{Evaluation on Backgrounds Challenge Dataset}
    \vspace{-10pt}
    \label{tab:bg_challenge}
\end{table*}

\begin{figure*}
    \centering
    \includegraphics[width = \linewidth]{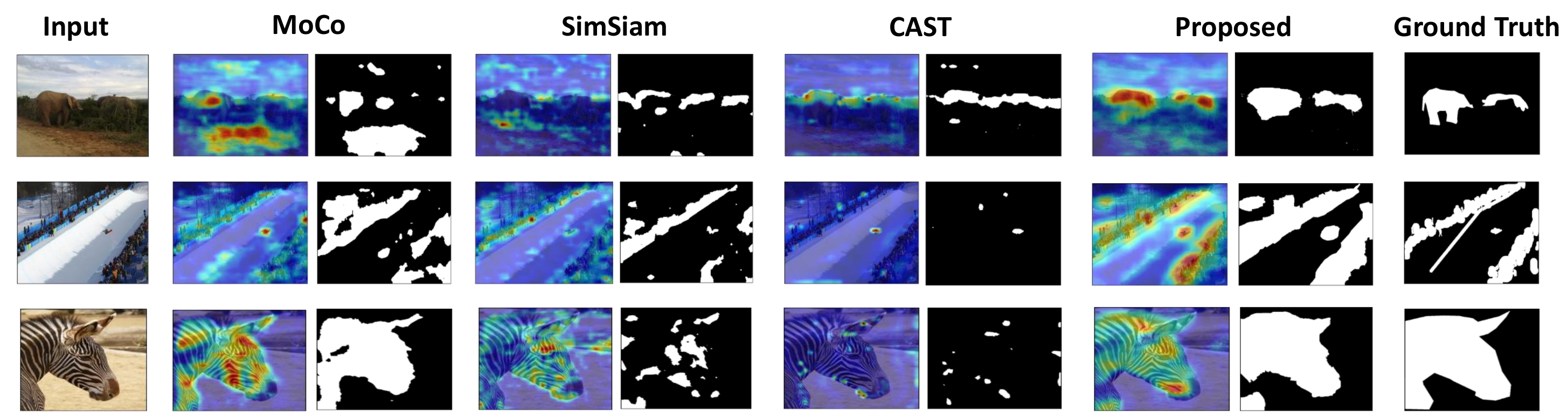}
    \caption{VDA maps of our proposed method are better priors for downstream segmentation with Grabcut. In each method's column, we show the VDA map prior and the resulting segmentation map.}

    \label{fig:grabcut_qual}
\end{figure*}

\subsection{Transfer Learning on Downstream Tasks}
Our first experiment follows the same protocol as CAST \cite{selvaraju2021casting}- we pretrain MoCo \cite{he2020momentum} with our proposed DiDA loss on images from the COCO dataset \cite{lin2014microsoft} and then evaluate the transfer performance on PASCAL VOC \cite{everingham2015pascal} and ImageNet \cite{russakovsky2015imagenet} classification, PASCAL VOC detection, and COCO \cite{lin2014microsoft} instance segmentation.
Full implementation details are in supp. material. 

\textbf{Classification:} We train a linear classification layer on top of our frozen pretrained model. For PASCAL VOC, we train using the VOC07 trainval split and report mAP numbers on the test split whereas for ImageNet-1K, we train on the ILSVRC 2012 train split and evaluate on the val split. 

\textbf{PASCAL VOC Object Detection:} We pretrain a ResNet-50-C4 \cite{he2016deep} Faster R-CNN \cite{ren2015faster,ren2016fasterpami} model with our DiDA loss, finetune using the trainval07+12 split, and evaluate on the test2007 split \cite{pascal-voc-2007}. 

\textbf{COCO Instance Segmentation:} We pretrain a Mask R-CNN \cite{he2017mask} model with the ResNet-50-FPN \cite{lin2017feature} backbone with our DiDA loss, finetune it using the trainval2017 split, and evaluate on the val2017 split. 

All results for these tasks are summarized in Table \ref{tab: results_downstream_quant}. In addition to a direct comparison with CAST \cite{selvaraju2021casting}, we also use these additional baselines: ``Random Init" uses a random initialization of model weights for task finetuning, ``MoCo-COCO" uses MoCo \cite{he2020momentum} weights for finetuning, and ``ImageNet Fully Sup" and ``COCO Fully Sup" represent fully supervised results. Like in CAST \cite{selvaraju2021casting}, the ``Random Init" and ``ImageNet Fully Sup" models are not evaluated for the classification task since this requires the convolutional neural network backbone to be frozen. We make several observations here. The proposed DiDA loss directly and substantially outperforms CAST on all downstream tasks with performance improvements of 0.8 mAP, 1.1 $AP^{bbox}_{75}$, and 1.3 $AP^{mask}_{75}$ for VOC07 classification, VOC detection, and COCO segmentation respectively. This result shows a model trained with our proposed DiDA loss is able to better capture all salient image regions. See also qualitative results in Figures~\ref{fig:teaser} and~\ref{fig:scene_attention_qual} where DiDA shows substantial improvements over CAST across multiple scenarios (e.g., row 4 in Figure~\ref{fig:scene_attention_qual} where DiDA highlights all animals much better and row 3 in Figure~\ref{fig:teaser} where DiDA highlights all person regions more clearly). Consequently, training with DiDA leads to a better initialization point, which in turn gives improved downstream results when compared to CAST. We show additional qualitative comparisons to CAST and more results on other datasets in the supplementary material.

Next, adding our DiDA loss to MoCo \cite{he2020momentum} helps substantially improve MoCO's performance (e.g., 67.5 mAP on VOC07 clf. vs. DiDA's 74.8, 51.5 $AP^{bbox}_{75}$ on detection vs. DiDA's 61.0, and 37.2 $AP^{mask}_{75}$ on segmentation vs. DiDA's 39.9). This suggests our proposed DiDA loss helps the baseline MoCo model learn better representations that further help with these transfer tasks. Note also substantial qualitative improvements in Figures~\ref{fig:teaser} and~\ref{fig:scene_attention_qual} where with DiDA, the model attends better to salient image regions when compared to the baseline MoCo model \cite{he2020momentum}. Finally, our overall numbers in some cases even outperform the fully supervised results, e.g., DiDA's 61.0 $AP^{bbox}_{75}$ on detection vs. the COCO fully supervised result of 54.7. Since the only difference between these two models is the starting point (random for COCO fully supervised and DiDA pretrained in our case), this suggests the proposed DiDA loss, by forcing the model to attend to all salient image regions, helps learn a stronger prior leading to downstream performance improvements.

\subsection{Evaluation on Backgrounds Challenge Dataset}

To evaluate how robust our proposed DiDA loss is to background variations, we use the backgrounds challenge dataset \cite{xiao2020noise} to conduct classification experiments. The idea here to evaluate our model on images where the foreground object of interest is placed in a variety of backgrounds (see Figure 1 in the supplementary material for some example images to get a sense of the variety in the background). In other words, if our model attended to the correct region of interest (foreground object), it would result in high accuracy despite background variations. To ensure a fair comparison with CAST \cite{selvaraju2021casting}, we use our model trained on the ImageNet-1K dataset from above and compute the classification accuracies shown in Table~\ref{tab:bg_challenge} on all the 8 foreground-background combinations in the dataset. One can note from the results that a model trained with our DiDA loss gives the highest accuracy across all combinations, 
providing additional evidence for DiDA's improved salient-region-grounding ability.

\subsection{Evaluating Visual Grounding}
\begin{figure}
    \centering
    \includegraphics[width = \linewidth]{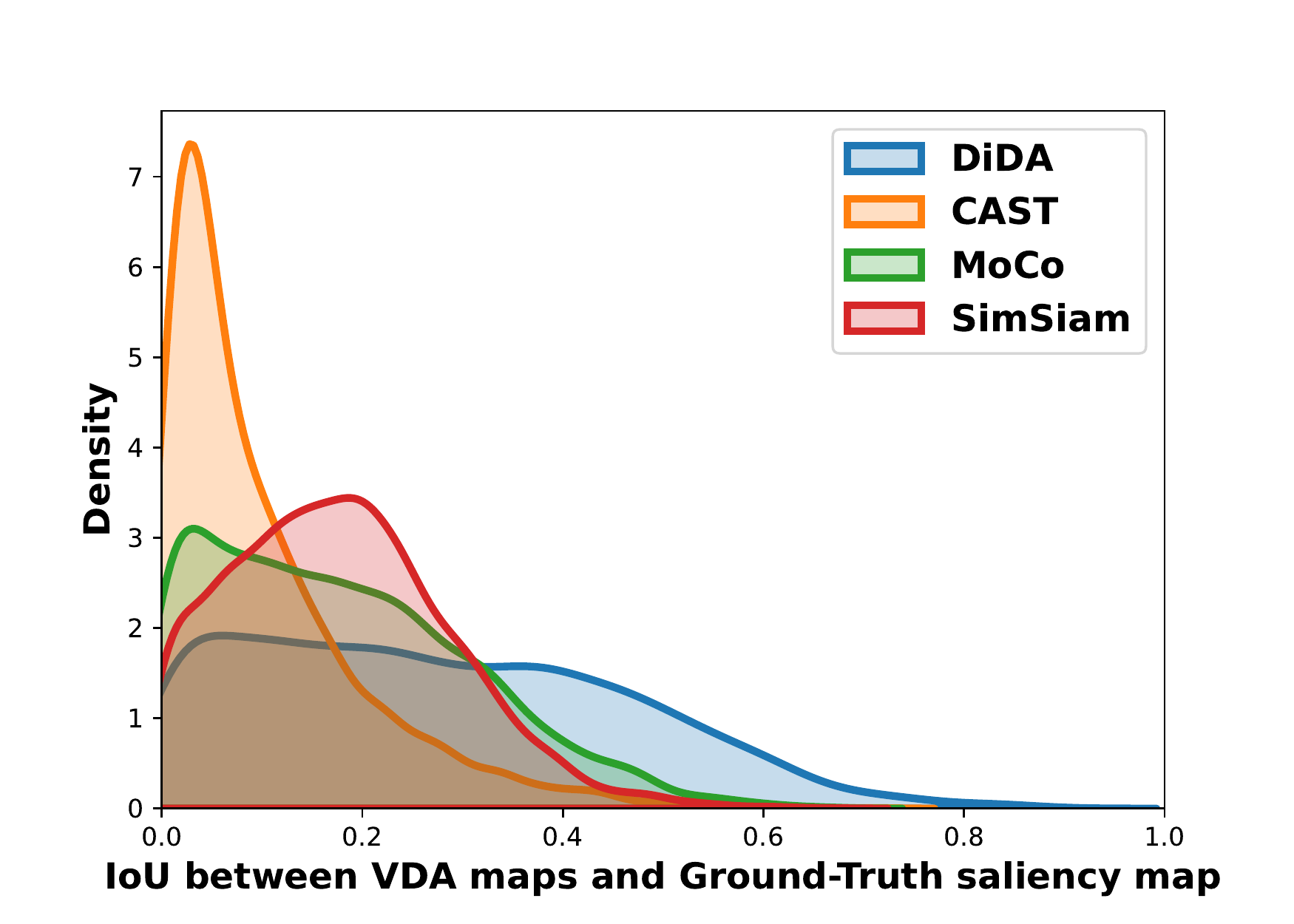}
    \vspace{-22pt}

    \caption{Visual Grounding Quantitative Analysis}
    \vspace{-1em}
    \label{fig:graph_iou_vg}
\end{figure}

In addition to Figures~\ref{fig:teaser} and~\ref{fig:scene_attention_qual}, we conduct additional experiments to provide more evidence for grounding. First, we quantify the improvement in visual grounding obtained using the proposed DiDA loss using the COCO val2017 split. To this end, we compute VDA maps using all models: MoCo \cite{he2020momentum}, SimSiam \cite{chen2020simple}, CAST \cite{selvaraju2021casting}, and our proposed method and then binarize them by thresholding at 0.5. We then compute the mean intersection-over-union (IoU) numbers for each method by comparing the VDA maps with the ground truth saliency map. The results are shown in Figure~\ref{fig:graph_iou_vg} using density estimation, where one can note substantial improvements for DiDA (the mIoU for DiDA is 0.29 when compared to 0.10 for CAST and 0.18 each for MoCo and SimSiam). This is due to our model's improved grounding to the foreground object. See supplementary material for more results.

In our next experiment, we evaluate how strong our attention maps can be as a prior for segmentation using interactive algorithms like GrabCut \cite{rother2004grabcut}. To this end, using data in the COCO val2017 split, we compute VDA attention maps using all models as above. We then use these attention maps as priors/starting points to run the GrabCut algorithm to produce segmentation maps, compare them to the ground truth, and compute the mIoU numbers. Specifically, we use the high- and low-response regions in the attention maps to label pixels as foreground/background prior to running GrabCut (see supplementary material for details). 
The results are shown in Table~\ref{tab:grabcut} where one can note substantial improvements with the proposed method (mIoU of 0.33) when compared to all other methods (mIoU of 0.25, 0.24, and 0.11 for MoCo, SimSiam, and CAST respectively). Since the attention maps with the model trained with our proposed DiDA loss has much better visual grounding to salient image regions when compared to other models, DiDA's attention maps provide a much better foreground/background starting point for the GrabCut algorithm, resulting in better final segmentation maps. To further substantiate this point, see Figure~\ref{fig:grabcut_qual} where one can note the attention maps of the proposed method localize salient image regions much better, resulting in better foreground and background seeds for GrabCut that then give better segmentation maps. See supplementary material for more results. 

\begin{table}
    \centering
    \scalebox{0.8}{
    \begin{tabular}{|c|c|c|c|c|}
    \hline
        \textbf{} & \textbf{MoCo} & \textbf{SimSiam} & \textbf{CAST} & \textbf{Proposed-DiDA} \\
        \hline
     mIoU &  0.25& 0.24& 0.11& 0.33 \\ 
     \hline
    \end{tabular}
    }
    \vspace{-0.5em}
    \caption{Mean IoU for GrabCut outputs.}
    \vspace{-2em}
    \label{tab:grabcut}
\end{table}

\section{Summary}
We considered the problem of self-supervised learning (SSL) of image representations. We first proposed a new concept called visual difference attention (VDA) that uses the feature vector difference between an image and its salient-region-masked-out variant to compute attention maps. Using them, we showed state-of-the-art SSL models have poor grounding to salient image regions. To address this limitation, we cast VDA as a differentiable operation and proposed a new loss called differentiable difference attention (DiDA). We showed training models with DiDA leads to both qualitatively better attention maps and quantitative improvements on tasks such as classification, detection, and segmentation. Additionally, with other proxy tasks like GrabCut segmentation, we showed the attention maps with our proposed loss can act as a strong scene prior for downstream segmentation, suggesting the model's ability to learn strong SSL representations.

{\small
\bibliographystyle{ieee_fullname}
\bibliography{egbib}

\begin{thebibliography}{10}\itemsep=-1pt

\bibitem{caron2018deep}
Mathilde Caron, Piotr Bojanowski, Armand Joulin, and Matthijs Douze.
\newblock Deep clustering for unsupervised learning of visual features.
\newblock In {\em Proceedings of the European conference on computer vision
  (ECCV)}, pages 132--149, 2018.

\bibitem{caron2019unsupervised}
Mathilde Caron, Piotr Bojanowski, Julien Mairal, and Armand Joulin.
\newblock Unsupervised pre-training of image features on non-curated data.
\newblock In {\em Proceedings of the IEEE/CVF International Conference on
  Computer Vision}, pages 2959--2968, 2019.

\bibitem{caron2020unsupervised}
Mathilde Caron, Ishan Misra, Julien Mairal, Priya Goyal, Piotr Bojanowski, and
  Armand Joulin.
\newblock Unsupervised learning of visual features by contrasting cluster
  assignments.
\newblock {\em Advances in Neural Information Processing Systems},
  33:9912--9924, 2020.

\bibitem{chattopadhay2018grad}
Aditya Chattopadhay, Anirban Sarkar, Prantik Howlader, and Vineeth~N
  Balasubramanian.
\newblock Grad-cam++: Generalized gradient-based visual explanations for deep
  convolutional networks.
\newblock In {\em 2018 IEEE winter conference on applications of computer
  vision (WACV)}, pages 839--847. IEEE, 2018.

\bibitem{chen2020adapting}
Lei Chen, Jianhui Chen, Hossein Hajimirsadeghi, and Greg Mori.
\newblock Adapting grad-cam for embedding networks.
\newblock In {\em Proceedings of the IEEE/CVF Winter Conference on Applications
  of Computer Vision}, pages 2794--2803, 2020.

\bibitem{chen2017deeplab}
Liang-Chieh Chen, George Papandreou, Iasonas Kokkinos, Kevin Murphy, and Alan~L
  Yuille.
\newblock Deeplab: Semantic image segmentation with deep convolutional nets,
  atrous convolution, and fully connected crfs.
\newblock {\em IEEE transactions on pattern analysis and machine intelligence},
  40(4):834--848, 2017.

\bibitem{chen2018encoder}
Liang-Chieh Chen, Yukun Zhu, George Papandreou, Florian Schroff, and Hartwig
  Adam.
\newblock Encoder-decoder with atrous separable convolution for semantic image
  segmentation.
\newblock In {\em Proceedings of the European conference on computer vision
  (ECCV)}, pages 801--818, 2018.

\bibitem{chen2020simple}
Ting Chen, Simon Kornblith, Mohammad Norouzi, and Geoffrey Hinton.
\newblock A simple framework for contrastive learning of visual
  representations.
\newblock In {\em International conference on machine learning}, pages
  1597--1607. PMLR, 2020.

\bibitem{chen2021exploring}
Xinlei Chen and Kaiming He.
\newblock Exploring simple siamese representation learning.
\newblock In {\em Proceedings of the IEEE/CVF Conference on Computer Vision and
  Pattern Recognition}, pages 15750--15758, 2021.

\bibitem{coates2012learning}
Adam Coates and Andrew~Y Ng.
\newblock Learning feature representations with k-means.
\newblock In {\em Neural networks: Tricks of the trade}, pages 561--580.
  Springer, 2012.

\bibitem{desai2021virtex}
Karan Desai and Justin Johnson.
\newblock Virtex: Learning visual representations from textual annotations.
\newblock In {\em Proceedings of the IEEE/CVF conference on computer vision and
  pattern recognition}, pages 11162--11173, 2021.

\bibitem{everingham2015pascal}
Mark Everingham, SM Eslami, Luc Van~Gool, Christopher~KI Williams, John Winn,
  and Andrew Zisserman.
\newblock The pascal visual object classes challenge: A retrospective.
\newblock {\em International journal of computer vision}, 111(1):98--136, 2015.

\bibitem{pascal-voc-2007}
M. Everingham, L. Van~Gool, C.~K.~I. Williams, J. Winn, and A. Zisserman.
\newblock The {PASCAL} {V}isual {O}bject {C}lasses {C}hallenge 2007 {(VOC2007)}
  {R}esults.
\newblock
  http://www.pascal-network.org/challenges/VOC/voc2007/workshop/index.html.

\bibitem{hadsell2006dimensionality}
Raia Hadsell, Sumit Chopra, and Yann LeCun.
\newblock Dimensionality reduction by learning an invariant mapping.
\newblock In {\em 2006 IEEE Computer Society Conference on Computer Vision and
  Pattern Recognition (CVPR'06)}, volume~2, pages 1735--1742. IEEE, 2006.

\bibitem{he2022masked}
Kaiming He, Xinlei Chen, Saining Xie, Yanghao Li, Piotr Doll{\'a}r, and Ross
  Girshick.
\newblock Masked autoencoders are scalable vision learners.
\newblock In {\em Proceedings of the IEEE/CVF Conference on Computer Vision and
  Pattern Recognition}, pages 16000--16009, 2022.

\bibitem{he2020momentum}
Kaiming He, Haoqi Fan, Yuxin Wu, Saining Xie, and Ross Girshick.
\newblock Momentum contrast for unsupervised visual representation learning.
\newblock In {\em Proceedings of the IEEE/CVF conference on computer vision and
  pattern recognition}, pages 9729--9738, 2020.

\bibitem{he2017mask}
Kaiming He, Georgia Gkioxari, Piotr Doll{\'a}r, and Ross Girshick.
\newblock Mask r-cnn.
\newblock In {\em Proceedings of the IEEE international conference on computer
  vision}, pages 2961--2969, 2017.

\bibitem{he2016deep}
Kaiming He, Xiangyu Zhang, Shaoqing Ren, and Jian Sun.
\newblock Deep residual learning for image recognition.
\newblock In {\em Proceedings of the IEEE conference on computer vision and
  pattern recognition}, pages 770--778, 2016.

\bibitem{jiang2021layercam}
Peng-Tao Jiang, Chang-Bin Zhang, Qibin Hou, Ming-Ming Cheng, and Yunchao Wei.
\newblock Layercam: Exploring hierarchical class activation maps for
  localization.
\newblock {\em IEEE Transactions on Image Processing}, 30:5875--5888, 2021.

\bibitem{kuznetsova2020open}
Alina Kuznetsova, Hassan Rom, Neil Alldrin, Jasper Uijlings, Ivan Krasin, Jordi
  Pont-Tuset, Shahab Kamali, Stefan Popov, Matteo Malloci, Alexander
  Kolesnikov, et~al.
\newblock The open images dataset v4: Unified image classification, object
  detection, and visual relationship detection at scale.
\newblock {\em International Journal of Computer Vision}, 128(7):1956--1981,
  2020.

\bibitem{lin2017feature}
Tsung-Yi Lin, Piotr Doll{\'a}r, Ross Girshick, Kaiming He, Bharath Hariharan,
  and Serge Belongie.
\newblock Feature pyramid networks for object detection.
\newblock In {\em Proceedings of the IEEE conference on computer vision and
  pattern recognition}, pages 2117--2125, 2017.

\bibitem{lin2014microsoft}
Tsung-Yi Lin, Michael Maire, Serge Belongie, James Hays, Pietro Perona, Deva
  Ramanan, Piotr Doll{\'a}r, and C~Lawrence Zitnick.
\newblock Microsoft coco: Common objects in context.
\newblock In {\em European conference on computer vision}, pages 740--755.
  Springer, 2014.

\bibitem{liu2020towards}
Wenqian Liu, Runze Li, Meng Zheng, Srikrishna Karanam, Ziyan Wu, Bir Bhanu,
  Richard~J Radke, and Octavia Camps.
\newblock Towards visually explaining variational autoencoders.
\newblock In {\em Proceedings of the IEEE/CVF Conference on Computer Vision and
  Pattern Recognition}, pages 8642--8651, 2020.

\bibitem{nguyen2019deepusps}
Tam Nguyen, Maximilian Dax, Chaithanya~Kumar Mummadi, Nhung Ngo, Thi
  Hoai~Phuong Nguyen, Zhongyu Lou, and Thomas Brox.
\newblock Deepusps: Deep robust unsupervised saliency prediction via
  self-supervision.
\newblock {\em Advances in Neural Information Processing Systems}, 32, 2019.

\bibitem{noroozi2016unsupervised}
Mehdi Noroozi and Paolo Favaro.
\newblock Unsupervised learning of visual representations by solving jigsaw
  puzzles.
\newblock In {\em European conference on computer vision}, pages 69--84.
  Springer, 2016.

\bibitem{pathak2016context}
Deepak Pathak, Philipp Krahenbuhl, Jeff Donahue, Trevor Darrell, and Alexei~A
  Efros.
\newblock Context encoders: Feature learning by inpainting.
\newblock In {\em Proceedings of the IEEE conference on computer vision and
  pattern recognition}, pages 2536--2544, 2016.

\bibitem{ren2015faster}
Shaoqing Ren, Kaiming He, Ross Girshick, and Jian Sun.
\newblock Faster r-cnn: Towards real-time object detection with region proposal
  networks.
\newblock {\em Advances in neural information processing systems}, 28, 2015.

\bibitem{ren2016fasterpami}
S Ren, K He, R Girshick, and J Sun.
\newblock Faster r-cnn: Towards real-time object detection with region proposal
  networks.
\newblock {\em IEEE Transactions on Pattern Analysis and Machine Intelligence},
  39(6):1137--1149, 2016.

\bibitem{rother2004grabcut}
Carsten Rother, Vladimir Kolmogorov, and Andrew Blake.
\newblock ``{G}rab{C}ut": interactive foreground extraction using iterated
  graph cuts.
\newblock {\em ACM transactions on graphics (TOG)}, 23(3):309--314, 2004.

\bibitem{russakovsky2015imagenet}
Olga Russakovsky, Jia Deng, Hao Su, Jonathan Krause, Sanjeev Satheesh, Sean Ma,
  Zhiheng Huang, Andrej Karpathy, Aditya Khosla, Michael Bernstein, et~al.
\newblock Imagenet large scale visual recognition challenge.
\newblock {\em International journal of computer vision}, 115(3):211--252,
  2015.

\bibitem{selvaraju2017grad}
Ramprasaath~R Selvaraju, Michael Cogswell, Abhishek Das, Ramakrishna Vedantam,
  Devi Parikh, and Dhruv Batra.
\newblock Grad-cam: Visual explanations from deep networks via gradient-based
  localization.
\newblock In {\em Proceedings of the IEEE international conference on computer
  vision}, pages 618--626, 2017.

\bibitem{selvaraju2021casting}
Ramprasaath~R Selvaraju, Karan Desai, Justin Johnson, and Nikhil Naik.
\newblock Casting your model: Learning to localize improves self-supervised
  representations.
\newblock In {\em Proceedings of the IEEE/CVF Conference on Computer Vision and
  Pattern Recognition}, pages 11058--11067, 2021.

\bibitem{tian2014learning}
Fei Tian, Bin Gao, Qing Cui, Enhong Chen, and Tie-Yan Liu.
\newblock Learning deep representations for graph clustering.
\newblock In {\em Proceedings of the AAAI Conference on Artificial
  Intelligence}, volume~28, 2014.

\bibitem{wang2020score}
Haofan Wang, Zifan Wang, Mengnan Du, Fan Yang, Zijian Zhang, Sirui Ding, Piotr
  Mardziel, and Xia Hu.
\newblock Score-cam: Score-weighted visual explanations for convolutional
  neural networks.
\newblock In {\em Proceedings of the IEEE/CVF conference on computer vision and
  pattern recognition workshops}, pages 24--25, 2020.

\bibitem{wei2022masked}
Chen Wei, Haoqi Fan, Saining Xie, Chao-Yuan Wu, Alan Yuille, and Christoph
  Feichtenhofer.
\newblock Masked feature prediction for self-supervised visual pre-training.
\newblock In {\em Proceedings of the IEEE/CVF Conference on Computer Vision and
  Pattern Recognition}, pages 14668--14678, 2022.

\bibitem{xiao2020noise}
Kai Xiao, Logan Engstrom, Andrew Ilyas, and Aleksander Madry.
\newblock Noise or signal: The role of image backgrounds in object recognition.
\newblock {\em arXiv preprint arXiv:2006.09994}, 2020.

\bibitem{yang2016joint}
Jianwei Yang, Devi Parikh, and Dhruv Batra.
\newblock Joint unsupervised learning of deep representations and image
  clusters.
\newblock In {\em Proceedings of the IEEE conference on computer vision and
  pattern recognition}, pages 5147--5156, 2016.

\bibitem{zhang2017split}
Richard Zhang, Phillip Isola, and Alexei~A Efros.
\newblock Split-brain autoencoders: Unsupervised learning by cross-channel
  prediction.
\newblock In {\em Proceedings of the IEEE conference on computer vision and
  pattern recognition}, pages 1058--1067, 2017.

\bibitem{ijcai2022p241}
Meng Zheng, Srikrishna Karanam, Terrence Chen, Richard~J. Radke, and Ziyan Wu.
\newblock Visual similarity attention.
\newblock In {\em Proceedings of the Thirty-First International Joint
  Conference on Artificial Intelligence, {IJCAI-22}}, pages 1728--1735, 2022.

\bibitem{zhou2016learning}
Bolei Zhou, Aditya Khosla, Agata Lapedriza, Aude Oliva, and Antonio Torralba.
\newblock Learning deep features for discriminative localization.
\newblock In {\em Proceedings of the IEEE conference on computer vision and
  pattern recognition}, pages 2921--2929, 2016.

\end{thebibliography}
}

\clearpage

\begin{appendices}

\section{}
\subsection{Background Challenge Dataset}

\begin{figure*}[h]
  \centering
  \includegraphics[width = 0.7\linewidth]{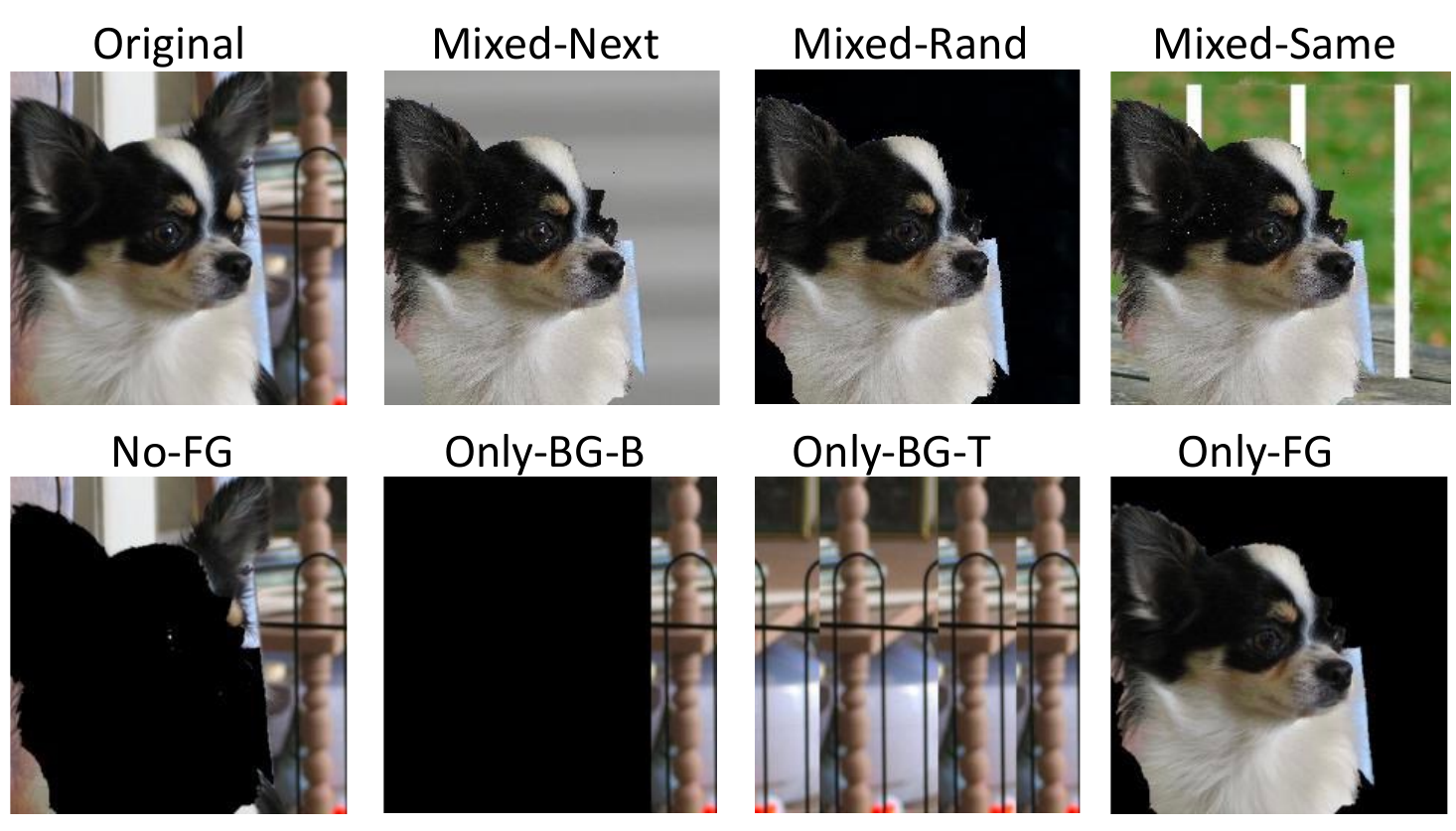}
  \caption{Sample images from background challenge dataset}
  \label{fig:bg_challenge_dataset}
\end{figure*}

The background challenge task (see Section 4.2 in main paper) seeks to evaluate the background-robustness of the models trained for the classification task. The idea here is to evaluate models in a variety of scenarios by placing the foreground object of interest on various different backgrounds. Figure~\ref{fig:bg_challenge_dataset} shows examples from the dataset for different foreground-background settings. 

\subsection{Implementation Details for Downstream Tasks}

\textbf{Classification:} We train a linear classification layer on top of our frozen pretrained model. For PASCAL VOC, we train using the VOC07 trainval split and report mAP numbers on the test split whereas for ImageNet-1K, we train on the ILSVRC 2012 train split and evaluate on the val split. For PASCAL VOC, we extract the $7 \times 7$ feature map from the last convolutional layer and downsample it to a $2 \times 2$ map via adaptive average pooling. We then flatten this feature map and normalize it to obtain 8192 dimensional features. Finally, we train SVMs for every class for costs $C \in$ {0.01, 0.1, 1.0, 10.0}, and select the best C by 3-fold cross validation. For ImageNet-1K, we extract a 2048-dimensional feature vector from the backbone network and train a linear layer on these features. We use SGD with momentum 0.9 and batch size 512 distributed across 8
Nvidia A100 GPUs. Like CAST, we follow VirTex \cite{desai2021virtex} for all other hyperparameter settings.

\textbf{PASCAL VOC Object Detection:} We pretrain a Faster R-CNN \cite{ren2015faster,ren2016fasterpami} model with the ResNet-50-C4 \cite{he2016deep} backbone with our DiDA loss, finetune it using the trainval07+12 split, and report accuracy metrics on the test2007 split \cite{pascal-voc-2007}. We train the model on 8 A100 GPUs for 24K iterations using SGD (0.9 momentum, 16 batch size, and $10^{-4}$ weight decay). Like in CAST, we start with a learning
rate of 0.02, use linear warmup for 100 iterations, and divide it by 10 at 18k and 22k iterations.

\textbf{COCO Instance Segmentation:} We pretrain a Mask R-CNN \cite{he2017mask} model with the ResNet-50-FPN \cite{lin2017feature} backbone with our DiDA loss, finetune it using the trainval2017 split, and report accuracy metrics on the val2017 split. We use a batch size of 16 and train the model on 8 A100 GPUs. Like in CAST, we use the $2 \times $ training schedule of Detectron2.

\subsection{Additional Qualitative Results}
\subsubsection{Downstream Segmentation with GrabCut}

We provide additional qualitative results in Figure~\ref{fig:grabcut_qual} to further support our claims that VDA maps obtained using our proposed method are better priors for segmentation using interactive methods such as GrabCut (see Secion 4.3 in main paper).  In each case, one can note the attention maps of the proposed method localize salient image regions much better (e.g., more high response coverage on the salient regions of interest). Such attention maps naturally lead to better foreground and background seeds for GrabCut, and consequently better segmentation maps as output. Please note these final segmentation results with our proposed method are closer to the ground truth when compared to other competing methods.

\textbf{\textit{Implementation Details:}}
We first compute attention maps using all the baseline methods and our proposed method. We then use these attention maps to obtain priors/starting masks to run the GrabCut algorithm. Specifically, we label pixels with high responses in the attention maps (values greater than 0.7) as \textsl{definitely foreground} and those with low response (values less than 0.1) as \textsl{definitely background}. Using such initialiazation, we run the GrabCut algorithm for all the methods and compute the final segmentation results.

\begin{figure*}
  \centering
  \includegraphics[width = \linewidth]{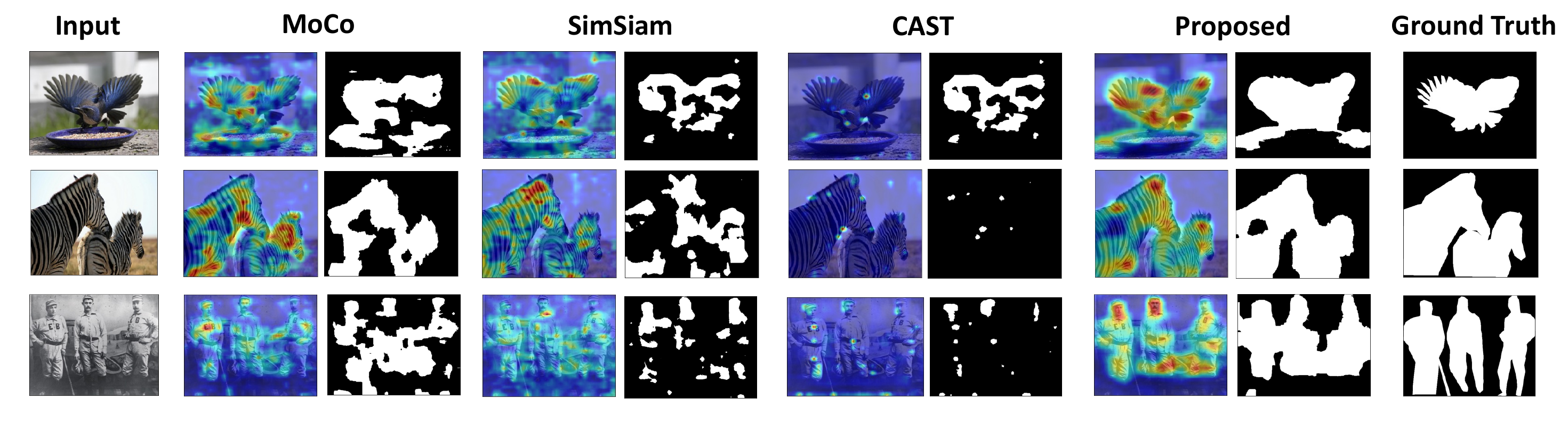}
  \caption{GrabCut qualitative results.}
  \label{fig:grabcut_qual}
\end{figure*}

\subsubsection{Additional attention maps}
\begin{figure*}
  \centering
  \includegraphics[width = 0.8\linewidth]{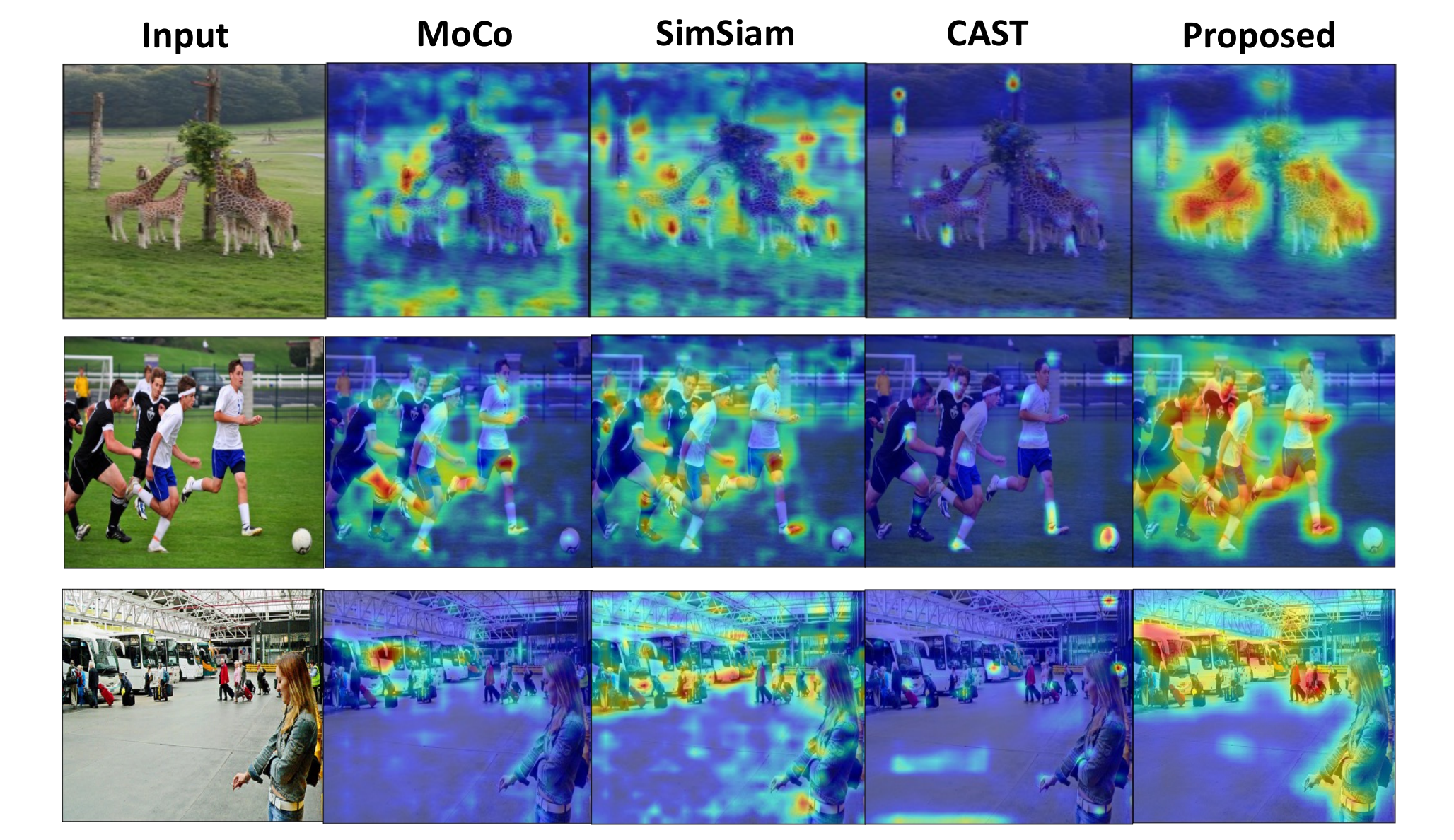}
  \caption{Additional attention map results of our proposed method vs. competing baselines.}
  \label{fig:scene_attention_qual_supp}
\end{figure*}

We present additional attention map results in Figure~\ref{fig:scene_attention_qual_supp} where one can see the visual grounding with our proposed method is substantially better when compared to existing methods. We show several complex scenarios in these examples where multiple objects are present in each image. Even in a fairly complex example such as the one shown in the third row, our proposed method is able to well attend to all the salient regions and outperforms all the baseline approaches. 

\begin{figure*}
  \centering
  \includegraphics[width = 0.8\linewidth]{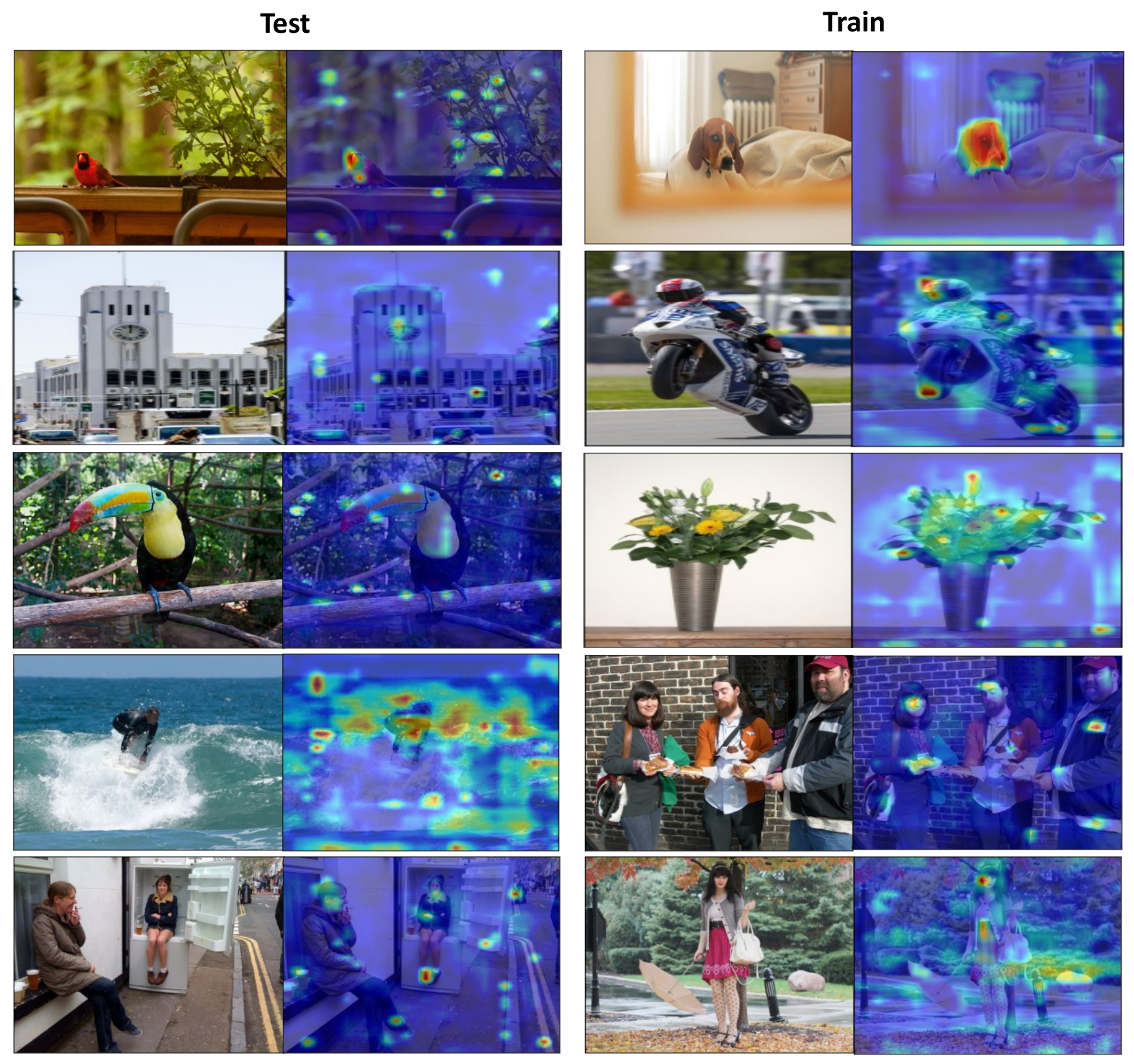}
  \caption{Additional attention map results for CAST on train and test images.}
  \label{fig:scene_attention_qual_cast}
\end{figure*}

Next, we show additional attention maps obtained with our closest competing method, CAST \cite{selvaraju2017grad}, in Figure~\ref{fig:scene_attention_qual_cast}. We show results on images from both train and test splits. As noted in Section 1 in the main paper, CAST has several limitations and they are reflected in these results. First, for training data, CAST shows good performance only in cases where there's a single salient object/region (e.g., the dog example in row 1 and flower vase example in row 3) whereas in cases that have more than one object, the attention maps are far from desirable. This is because the CAST loss compares crop attention (computed on a crop from the full image) to a full-image saliency map and by design this does not highlight all salient regions. Due to such suboptimal supervision, results on testing data are also not as desired, with the model attention clearly not focused on the salient aspects in these cases. 

\begin{figure*}
  \centering
  \includegraphics[width = \linewidth]{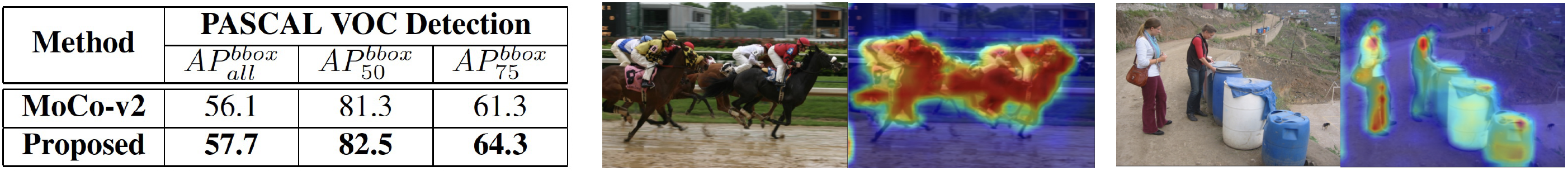}
  \caption{Additional results for the proposed method when pre-trained on OpenImages dataset.}
  \label{fig:results_ohms}
\end{figure*}

We further show the effectiveness of DiDA by providing results for a model trained with our DiDA loss on OpenImages dataset \cite{kuznetsova2020open}. Please see the Table (we outperform MoCo-v2 after pretraining with OpenImages) and the attention maps in Figure \ref{fig:results_ohms}.

\subsection{Visual Difference Attention}
\begin{figure*}
  \centering
  \includegraphics[width = \linewidth]{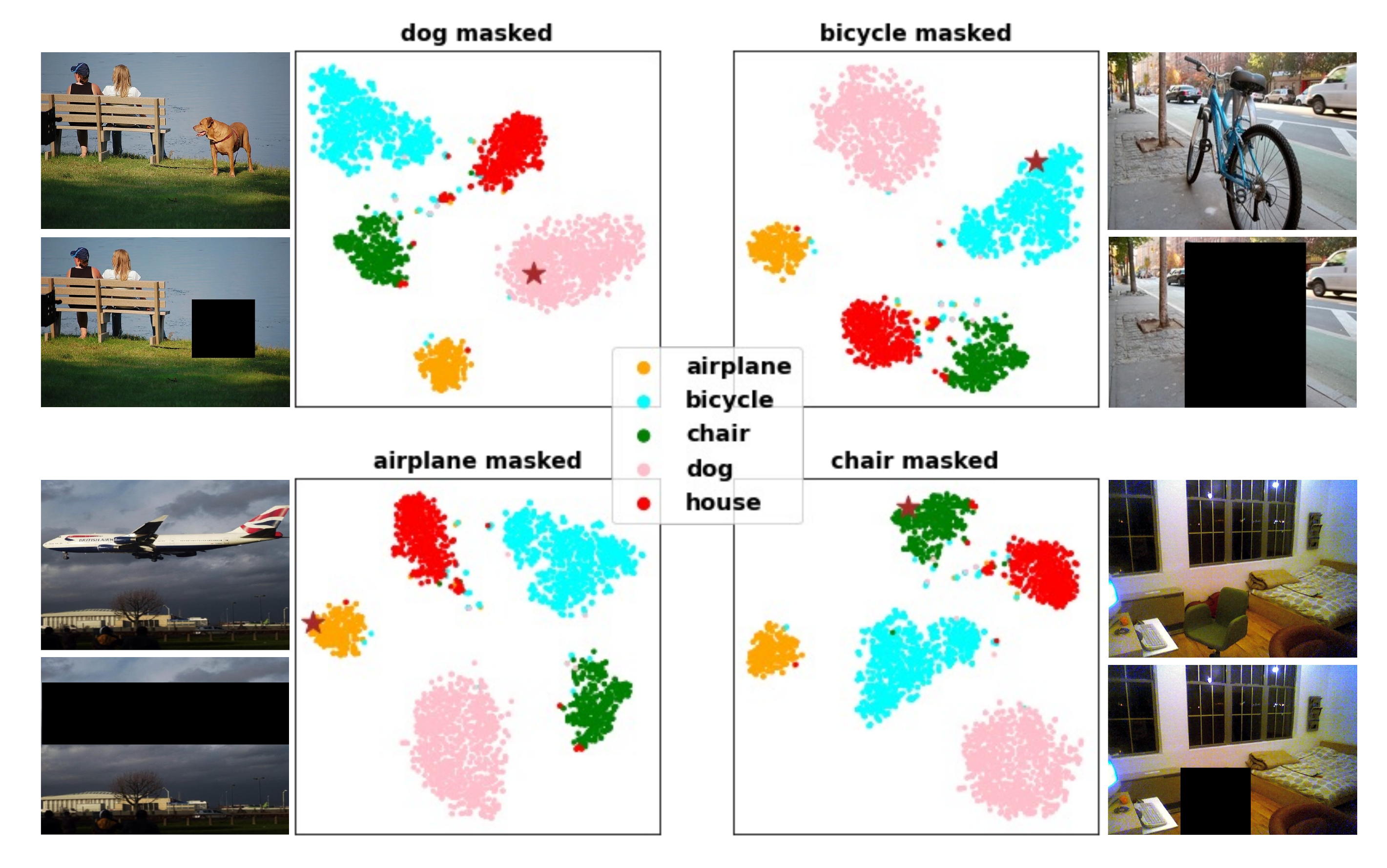}
  \caption{t-SNE plot of Figure 3 in the main paper with the original and masked images shown in each case.}
  \label{fig:tsne_vis_diff}
\end{figure*}

In Section 3.1 in the main paper, we proposed the idea of masking out a salient region in an image and computing the difference between the resulting and the original image to determine the importance an SSL model assigns to this region. To substantiate this intuition, we showed results of a controlled experiment in Figure 3 (in the main paper). Here, we show the actual images used in this experiment. In Figure~\ref{fig:tsne_vis_diff}(a), we compute the difference feature between the original image and the dog-masked image and show the resulting difference vector falls into the dog cluster in the t-SNE plot (as expected). Similarly, in Figure~\ref{fig:tsne_vis_diff}(b), we obtain the difference feature vector from the bicycle image and the bicycle-masked image and show it falls into the bicycle cluster in the t-SNE plot as expected.

\end{appendices}
\end{document}